\documentclass[letterpaper, 10 pt, conference]{ieeeconf}  

\IEEEoverridecommandlockouts                              

\usepackage{graphicx}        
\usepackage{multicol}        
\usepackage[bottom]{footmisc}
\usepackage[capbesideposition=outside,capbesidesep=quad]{floatrow}
\usepackage{hyperref}
\usepackage[export]{adjustbox}
\usepackage{cite}
\hypersetup{bookmarksopen,bookmarksnumbered,
pdfpagemode=UseOutlines,
colorlinks=true,
linkcolor=blue,
anchorcolor=blue,
citecolor=blue,
filecolor=blue,
menucolor=blue,
urlcolor=blue
}
\usepackage{marvosym}

\usepackage[T1]{fontenc}    
\usepackage{upgreek}
\usepackage{amsfonts,amssymb}
\usepackage{bm}
\usepackage{bbm}

\usepackage{amsthm} 
\usepackage{thmtools}
\usepackage{mathtools}
\usepackage{epstopdf}
\usepackage{xspace}
\usepackage{outlines}
\usepackage[]{caption} 
\usepackage[font=footnotesize]{subcaption}
\usepackage{units}
\usepackage{colortbl}
\usepackage{tabulary}
\usepackage[usenames,dvipsnames]{xcolor} 
\usepackage{diagbox}
\usepackage[ruled,vlined,linesnumbered]{algorithm2e}  
\usepackage{parskip}
\SetKwComment{Comment}{$\triangleright$\ }{}
\usepackage{ifthen,version}
\usepackage{soul}
\usepackage{fixltx2e}
\usepackage{csquotes}
\usepackage{microtype}      
\usepackage{lipsum}
\usepackage{csquotes}
\usepackage{tikz}
\let\labelindent\relax
\usepackage{enumitem}
\usepackage{wrapfig}
\usepackage{siunitx}
\usepackage[normalem]{ulem}

\usepackage{multirow}
\usepackage{pbox}
\usepackage{blindtext}
\usepackage{duckuments}

\SetCommentSty{mycommfont}

\SetKwInput{KwInput}{Input}                
\SetKwInput{KwOutput}{Output}              

\newtheoremstyle{hypstyle}
{3pt} 
{3pt} 
{\itshape} 
{} 
{\bfseries} 
{.} 
{.5em} 
{} 

\theoremstyle{hypstyle}

\DeclareMathOperator*{\argmin}{arg\,min}

\newcommand{\argminprob}[1]{\underset{#1}{\argmin}}

\newcommand{\real}[0]{\mathbb{R}}

\newcommand{\bbm}{\begin{bmatrix}}
\newcommand{\ebm}{\end{bmatrix}}

\newcommand{\Path}[1]{\xi_{#1}}
\newcommand{\cost}[0]{J}
\newcommand{\costFn}[1]{\cost \left( #1 \right)}

\newcommand{\costSmooth}[0]{\cost_\mathrm{smooth}}
\newcommand{\costFnSmooth}[1]{\costSmooth \left( #1 \right)}

\newcommand{\costOcc}[0]{\cost_\mathrm{occlusion}}
\newcommand{\costFnOcc}[1]{\costOcc \left( #1 \right)}

\newcommand{\costObs}[0]{\cost_\mathrm{obstacle}}
\newcommand{\costFnObs}[1]{\costObs \left( #1 \right)}

\newcommand{\costForm}[0]{\cost_\mathrm{form}}
\newcommand{\costFnForm}[1]{\costForm \left( #1 \right)}

\newcommand{\map}[0]{\mathcal{M}}
\newcommand{\grid}[0]{\mathcal{G}}

\newcommand{\hpose}[0]{\mathcal{\theta}}

\newcommand{\Traj}[0]{\xi}

\graphicspath{{figs/}}
\usepackage{balance}

\title{
\LARGE\bf3D Human Reconstruction in the Wild \\ with Collaborative Aerial Cameras 
\vspace{-2mm}
}

\author{Cherie Ho$^{1}$, Andrew Jong$^{1}$, Harry Freeman$^{1}$, Rohan Rao$^{1}$, Rogerio Bonatti$^{1}$, Sebastian Scherer$^{1}$ 
\vspace{-1.36mm}
\thanks{$^{1}$The Robotics Institute, Carnegie Mellon University, Pittsburgh PA
        {\tt\small \{cherieh, ajong, hfreeman, rgrao, rbonatti, basti\}@cs.cmu.edu}}%
}

\let\oldtwocolumn\twocolumn
\renewcommand\twocolumn[1][]{%
    \oldtwocolumn[{#1}{
    \begin{center}
           \includegraphics[width=1.0\textwidth]{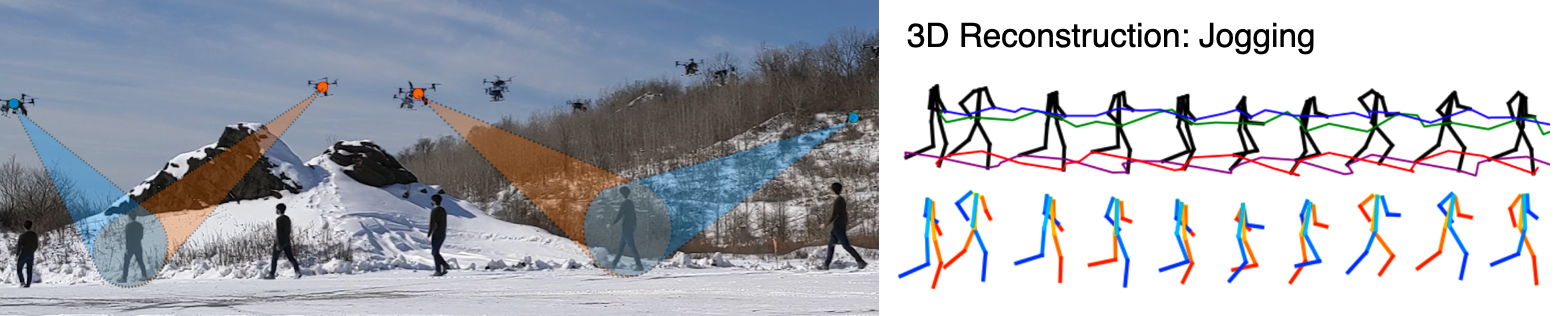}
           \captionof{figure}{\small We present a multi-UAV system for 3D human reconstruction in the wild.
           Our framework coordinates the motion of multiple aerial cameras to optimally reconstruct the dynamic target's 3D body pose while avoiding obstacles and occlusions.
           We deploy the system in challenging real-world conditions and target motions such as jogging and playing soccer. 
    	   \vspace{2.0mm}
    	   }
           \label{fig:main}
        \end{center}
    }]
}

\begin{document}

\maketitle
\thispagestyle{empty}
\pagestyle{empty}


\begin{abstract}

Aerial vehicles are revolutionizing applications that require capturing the 3D structure of dynamic targets in the wild, such as sports, medicine and entertainment.
The core challenges in developing a motion-capture system that operates in outdoors environments are: 
(1) 3D inference requires multiple simultaneous viewpoints of the target,
(2) occlusion caused by obstacles is frequent when tracking moving targets, 
and (3) the camera and vehicle state estimation is noisy.
We present a real-time aerial system for multi-camera control that can reconstruct human motions in natural environments without the use of special-purpose markers.
We develop a multi-robot coordination scheme that maintains the optimal flight formation for target reconstruction quality amongst obstacles. We provide studies evaluating system performance in simulation, and validate real-world performance using two drones while a target performs activities such as jogging and playing soccer.\\
Supplementary video: \url{https://youtu.be/jxt91vx0cns}

\end{abstract}

\section{Introduction}
\label{sec:intro}

3D reconstruction of scenes with stereo cameras, RGB-D sensors and monocular cameras is a topic intensively studied in the computer vision and robotics communities~\cite{izadi2011kinectfusion, geiger2011stereoscan, engel2014lsd}. However, most works focus on static environments, and can only reconstruct static objects. Until recently, capturing dynamic scenes could only be achieved using body markers and high-precision motion capture systems \cite{chan2011virtual, fern2012biomechanical}, or by markerless systems that heavily rely on skeleton models \cite{gall2009motion}. Pan-optic studios, on the other hand, rely on visual data from a large number of static cameras to precisely capture motions of multiple targets \cite{kanade2001large,kanade2001sports,joo2015panoptic, simon2017hand, zhang2018multiview}. However, they require expensive structures and are confined to small indoor areas.

Aerial camera technologies can significantly extend the capabilities of recording setups to handle dynamic targets in natural outdoor environments.
Several works allow drones to detect, track and follow targets in real-time using single-camera systems~\cite{bonatti2020autonomous,bonatti2019towards,gschwindt2019can,bonatti2018autonomous,gebhardt2016airways, gebhardt2018optimizing, joubert2015interactive, roberts2016generating}.
We also find a rich history of work on multi-camera aerial systems that allow users to estimate poses of moving targets.
For instance, \cite{Nageli2018flycon} uses multiple drones to capture the pose of a human wearing markers. \cite{tallamraju2020aircaprl} explicitly optimizes for body pose reconstruction, but in obstacle-free environments. 
Most related to our work, \cite{saini2019aircap} presents an aerial motion capture system that uses multi-robot formation controller \cite{tallamraju2019active} for data collection. The multi-robot controller avoids obstacles while maintaining formation, but it plans for a short time horizon (1.5s). 

Despite the recent progress in multi-drone recording systems, existing approaches are still not able to simultaneously handle all major challenges related to 3D dynamic pose reconstruction in natural environments without markers. A robust system for 3D inference must be able to coordinate the simultaneous recording of various viewpoints of the target even within the presence of obstacles, which can cause image occlusions and robot collisions. In addition, the robots need to navigate smoothly to avoid state estimation noise. Finally, when filming a dynamic target in complex environment, the robot needs to anticipate future actor motions and compute long horizon plans for more optimal motions.

As seen in Figure~\ref{fig:main}, we tackle these challenges by building upon our previous work in multi-drone cinematography \cite{bucker2020}, to which a new formation control strategy is introduced for the human reconstruction problem. The proposed system can plan over long horizons multi-robot trajectories for human reconstruction while avoiding occlusions and obstacles. 
Our contributions are three-fold:

\textbf{1) Multi-camera coordination:}  We formulate a multi-camera coordination scheme with the goal of maximizing the reconstructed 3D pose quality of dynamic targets. We develop a scalable two-stage system with long planning time horizons and real-time performance that uses a centralized planner for formation control and a decentralized trajectory optimizer that runs on each robot;

\textbf{2) System scaling and error ablations:} We provide extensive simulation experiments validating system performance under different operating conditions such as scaling over multiple drones, and reconstruction error analysis for different magnitudes of camera pose uncertainty.

\textbf{3) Real-world experiments: } We deploy the system in real-world settings using two robots while tracking an actor performing activities such as jogging and playing soccer. We empirically show the improvement in reconstruction quality caused by our adaptive formation scheme.


\section{Related Work}
\label{sec:related_work}

\textbf{Aerial Cinematography and Active Tracking:}
There is a significant body of work in academia and in industry within the domain of single-drone cinematography. For instance, \cite{bonatti2018autonomous, bonatti2019towards, bonatti2020autonomous} compute smooth aerial camera plans given user-defined artistic guidelines. As well, commercial drones from Skydio \cite{skydio} and DJI \cite{mavic} can track and film actors in complex cluttered environments. Recently, there is growing interest in coordinating multiple UAVs to add viewpoint diversity, with pioneering work that proposes online path planning with inter-drone collision avoidance for indoor settings \cite{nageli2017real}.
\cite{TorresGonzlez2019DistributedME, Alcntara2020AutonomousEO, Alcntara2020OptimalTP, caraballo2020autonomous, Karakostas2020ShotTC} focus on coordinating multi-drone, human-guided shot execution under various constraints, such as smoothness, battery life, and mutual camera visibility.
Our previous work \cite{bucker2020} increases practicality for multi-UAV tracking of unscripted targets by removing the need for predefined shots while maximizing 3D shot diversity online. 

\textbf{Aerial Motion Capture:}
Traditionally, human motion capture is achieved by tagging targets with body markers, where recent progress lies in accurate  markerless reconstruction using a large number of static visual cameras \cite{joo2015panoptic}. 
Subsequent works investigate mobile UAVs as an alternative to overcome the complexity of this static setup, with \cite{pirinen2019domedrone, zhou2018mocap, kiciroglu2019activemocap} focusing on optimal camera plans for a single UAV. We find works  that plan optimal viewpoints with multiple vehicles \cite{xu2018flycap, tallamraju2020aircaprl}; however, they do not consider obstacle avoidance. Most related to our work, \cite{tallamraju2019active} introduces a decentralized multi-UAV coordination framework for actor position estimation that is extended as a data collection system for outdoor human shape estimation \cite{saini2019aircap}. The system plans for a short horizon (1.5s) to avoid obstacles while maintaining formation around target. However, in our previous work \cite{bonatti2020autonomous}, we have shown that planning with longer horizons minimizes the likelihood of myopic trajectories for better target tracking in complex environments.
In this work, we present a multi-UAV motion capture system that plans smooth trajectories over a long time horizon (10s) to maintain optimal flight formation for target reconstruction among obstacles. 


\section{Problem Definition}
\label{sec:prob_def}

Our overall goal is to control a team of aerial cameras to reconstruct the 3D pose of a dynamic human moving through a cluttered environment.
Let $\hpose(t) \in \real^{P \times 3}$ be a vector containing the target's 3D coordinates for $P$ joints at time $t$.
Our mathematical objective is to minimize the reconstruction error $E_{\text{recon}}$ calculated with respect to the true target joints over time:
\vspace{-6mm}

\begin{equation}
E_{\text{recon}} = \sum_{t=1}^{T} || \hat{\hpose}(t) - \hpose(t)  ||^2
\label{eq:reconstruction_loss}
\end{equation}

\vspace{-5mm}
In order to minimize this objective (Eq. \ref{eq:reconstruction_loss}), we capture the scene using a set of aerial cameras, and calculate their trajectories using an optimization framework.
Similarly to \cite{bonatti2020autonomous}, we employ a weighted set of cost functions 
that balance robot safety and motion smoothness against visual occlusions of the actor from obstacles.
In addition, we introduce a new objective to encode multi-camera collaboration that strives to keep an optimal drone formation over time.

Let $\Path{qi} : [0,t_f] \rightarrow \real^3  \times SO(2)$ be the trajectory of the \textit{i}-th UAV, i.e., $\Path{qi}(t) = \{x(t), y(t), z(t), \psi_{q}(t)\}$, and $\Xi = \{\Path{q1}, ..., \Path{qn} \}$ be the set of trajectories from $n$ UAVs. Let $\Path{a} : [0,t_f] \rightarrow \real^3$ be the trajectory of the actor, i.e., $\Path{a}(t) = \{x(t), y(t), z(t)\}$, which is inferred using onboard cameras. Let grid $\grid: \real^3 \rightarrow \real$ be a voxel occupancy grid that maps every point in space to a probability of occupancy. Let $\map(\grid) : \real^3 \rightarrow \real$ be the signed distance values of a point to the nearest obstacle. 

Each objective is represented as follows:

\textit{1) Smoothness:} Penalizes jerky motions that may lead to camera blur and unstable flight. Calculated as the sum of costs from individual trajectories:
$\costFnSmooth{\Xi} = \sum_i \costFnSmooth{\Path{qi}}$;\\
\textit{2) Occlusion:} Penalizes occlusion of the actor by obstacles in the environment for each camera:
$\costFnOcc{\Xi} = \sum_i \costFnOcc{\Path{qi},\Path{a},\map}$;\\
\textit{3) Obstacle:} Penalizes proximity to obstacles that are unsafe for each UAV:
$\costFnObs{\Xi} = \sum_i \costFnObs{\Path{qi},\map}$;\\
\textit{4) Formation:} Ensures that the camera formation remains at the optimal configuration for actor reconstruction. Calculated over the entire set of trajectories:
$\costFnForm{\Xi,\Path{a}}.$

We then compose the overall cost function as a linear combination between each component, with relative weights $\lambda$. 

The solution $\Xi^*$ is then tracked by each UAV:

\begin{equation}
\begin{aligned}
\label{eq:main_cost}
\costFn{\Xi} &=  \begin{bmatrix}
       1 & \lambda_1 & \lambda_2 & \lambda_3 
     \end{bmatrix} 
\begin{bmatrix}
           \costFnSmooth{\Xi} \\
           \costFnOcc{\Xi} \\
           \costFnObs{\Xi} \\
           \costFnForm{\Xi}
     \end{bmatrix} \\
\Xi^* &= \argminprob{} \quad \costFn{\Xi}
\end{aligned}
\end{equation}


\section{Approach} 
\label{sec:approach}

We now detail the methods we use for camera coordination in the multi-UAV system. 
As displayed in Equation~\ref{eq:reconstruction_loss}, our overall objective function involves the minimization of 4 sub-objectives, which may often conflict with one another. 
Our goal is to formulate an algorithm that works in real time in unscripted scenes, and scales to more UAVs without a large computational penalty.

To address the time complexity issue, we break down our method into three main subsystems operate together.
First, a centralized motion planner (Sec. \ref{subsec:centralized}) coordinates desired positions for all cameras simultaneously. 
Next, on each UAV, a decentralized motion planner (Sec. \ref{subsec:decentralized}) computes the final trajectories for the specific UAV. 
Finally, an offline skeletal reconstruction module (Sec. \ref{subsec:reconstruction}) processes  images from all cameras to output the actor's pose vector over time. 
Fig. 2 depicts the system diagram.

\begin{figure}[t]
  \includegraphics[width=\linewidth]{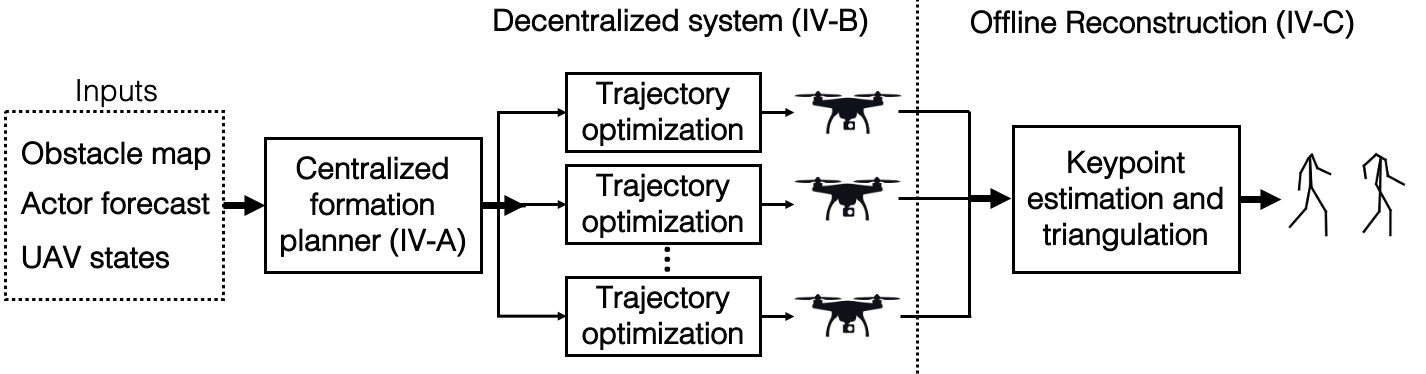}

   \caption{\small System overview: a centralized formation planner computes discrete camera positions for optimal target reconstruction. Next, each UAV optimizes smooth trajectories to follow. Finally, the collected multi-view footage and camera poses are used offline to reconstruct a sequence of the target's joints in 3D.} 

     \label{fig:approach}
\end{figure}

\subsection{Centralized Formation Planning}
\label{subsec:centralized}

\begin{figure}[H]
   \centering
   \includegraphics[width=0.85\textwidth]{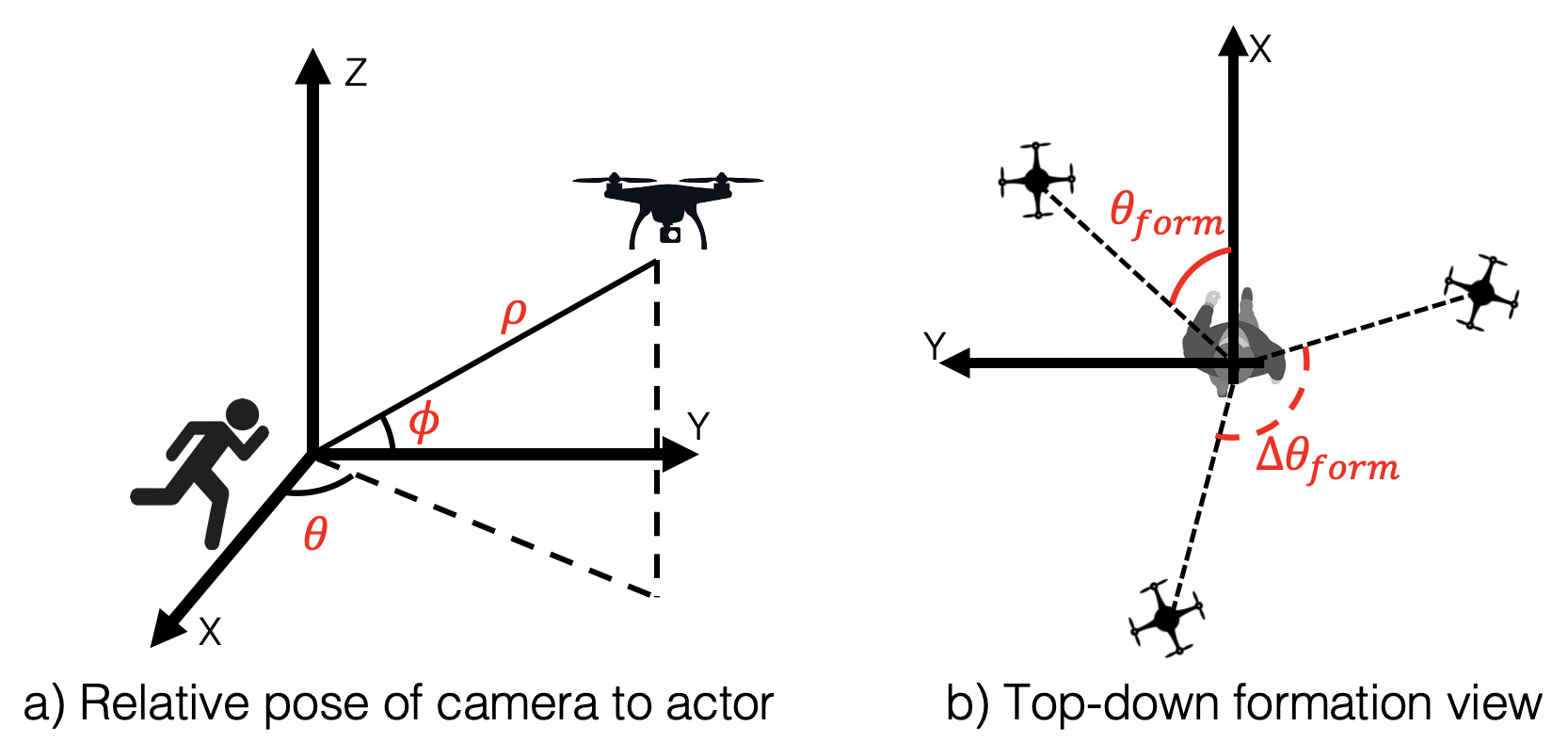}
   \caption{\small (a) Spherical camera coordinates relative to actor. (b) Top-down view of formation showing formation yaw $\theta_{form}$ and desired yaw angle difference $\Delta\theta_{form}$.
   \vspace{-2.0mm}}
   \label{fig:coord-fig}
\end{figure}

Our centralized formation planning system parametrizes trajectories as waypoints, i.e. $\Traj \in \real^{T\times3}$, where $T$ is the number of time steps.
Actor trajectory $\Path{a}$ is forecasted given current actor pose using a Kalman Filter with a constant-velocity model.
We assume the UAV heading direction $\psi(t)$ is set to always point the drone from $\Traj_{qi}(t)$ towards the actor in $\Traj_a(t)$, which can be achieved independently of the aircraft's translation by rotating the UAV's body and camera gimbal. 
We extend our previous  multi-drone cinematography work \cite{bucker2020} and plan formation trajectories using a state-space parametrized in spherical coordinates $\{\rho, \theta, \phi\}$ centered on the actor's position (Fig. \ref{fig:coord-fig}a): 

\begin{equation}
  \begin{aligned}
    &\Path{qi}(t) = \Path{a}(t) + \rho 
    \begin{bmatrix}
     cos(\theta_i) cos(\phi_i)\\
     sin(\theta_i) cos(\phi_i)\\
     sin(\phi_i)
    \end{bmatrix}\\
  \end{aligned}
\end{equation}

Next, we mathematically formulate the cost functions that the centralized formation planner optimizes for the formation trajectory set $\Xi$:

\textit{i) Formation:} To minimize the pose reconstruction error (Eq. \ref{eq:reconstruction_loss}), the ideal camera formation should maximize the angular distance between the multiple cameras.
We define the optimal formation for $n$ UAVs (Fig. \ref{fig:coord-fig}b) as points with equidistant yaw angles relative to the target, where $\Delta\theta_{\text{form}} = \frac{2\pi}{n}$, and with a special case of $\Delta\theta_{\text{form}} = \frac{\pi}{2}$ for $n=2$.
The desired tilt angle $\phi_{\text{form}}$ and radius $\rho_{\text{form}}$ are equal for all UAVs. The cost is calculated as:
\vspace{-5mm}

\begin{equation}
   \begin{aligned}
   \label{eq:cost_formation}
       \costFnForm{\Xi} = &\sum_{t=1}^{T} \sum_{i=1}^{n} || \Traj_{i}(t) - \Traj_{i\ \text{form}}(t) ||
   \end{aligned}
\end{equation}

\begin{wrapfigure}{R}{0.4\linewidth}
   \centering
  \vspace{-7mm}
   \includegraphics[width=1.0\linewidth]{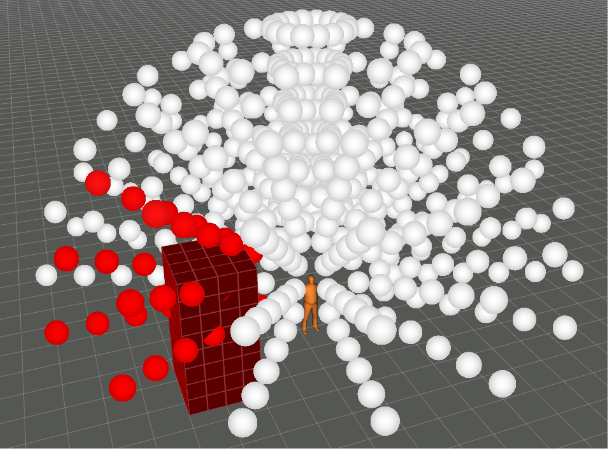}
  \vspace{-7mm}
   \caption{\small Visualization of occupancy and occlusion avoidance costs in spherical grid $\grid_{s}^{t}$, from \cite{bucker2020}.
   \vspace{-2mm}}
   \label{fig:state_space}
\end{wrapfigure}

\textit{ii) Safety: }
To maintain safety, we must reason about the role of obstacles in the environment.
First, we transform the environment's occupancy grid into a time-dependent spherical domain centered around the target $\grid \rightarrow \grid_{s}^{t} \in [0,1]$, as shown in Fig.~\ref{fig:state_space} and Eq.~\ref{eq:cost_obs}. 
\vspace{-2mm}

\begin{equation}
\begin{aligned}
\label{eq:cost_obs}
    \costFnObs{\Xi} &= \sum_{t=1}^{T} \sum_{i=1}^{n} \int_0^{r_{max}} \grid_{s}^{t}({\Path{qi}(t)}) \ d(\text{volume})
\end{aligned}
\end{equation}

\vspace{-2mm}
\textit{iii) Occlusion avoidance: }
In order to maintain target visibility at all times, we calculate the occlusion cost as a measure of occupancy along a line $l_i(\tau) = \tau\Path{qi}(t) + (1-\tau)\Path{a}(t)$ between UAV and target:
\vspace{-5mm}

\begin{equation}
\begin{aligned}
\label{eq:cost_occ}
    \costFnOcc{\Xi} &= \sum_{t=1}^{T} \sum_{i=1}^{n} \int_0^{1}\grid_{s}^{t}({l_i(\tau)}) \ d\tau
\end{aligned}
\end{equation}

\vspace{-2mm}
Next, we find the optimal sequence of angles for the UAV formation that minimizes the sum of costs.
Instead of solving for each UAV path sequentially as in our previous work \cite{bucker2020}, in this work we assume a fixed drone formation, and only find the optimal yaw angle sequence over $T$ time steps: $\Theta_{form}^* = \{ \theta_1, ..., \theta_T \}$.

We define a state space $S$ with all possible formation yaw values, where $|S| = T \times D$, where $D$ is the number of discrete values over the interval $[-\pi, \pi]$.  
We build a cost map $C: S \rightarrow \real^{|S|}$ that contains the cost of all states, and a cost-to-go map $V: S \rightarrow \real^{|S|}$.
In order to make transitions between cells dynamically feasible for the real vehicle, we only allow expansions to neighboring cells in the next ring.
Given that we operate in a discrete state-space with a relatively small branching factor and deterministic transitions, a single backwards dynamic programming pass yields the optimal solution in little time.
Finally, we build the full formation yaw sequence $\Theta_{form}^*$ by selecting neighboring cells with the least cost-to-go at consecutive time steps, starting at the formation's initial yaw $\theta_0$.
Algorithm~\ref{alg:formation_traj} details the process.

\begin{algorithm}[ht]
   \small
\caption{\small Compute formation $\Theta_{form}^* = \{ \theta_1, ..., \theta_T \}$}
\label{alg:formation_traj}
\SetAlgoLined
  $C$ $\gets$ $J(S)$; \Comment{\footnotesize update formation cost map}
  $V_T$ $\gets$ $C$; \Comment{\scriptsize initialize cost-to-go at time T}
  \Comment{\footnotesize Begin backwards pass}
  \For{$t = T-1, T-2, ..., 1$}  { 
   \For{$i = 1, ..., D$}{
    $V_t^i$ $\gets$ $\argminprob{i} V_{t+1}^i$ + $C^i$; \Comment{\scriptsize neighbors}
   }
  }
  \Comment{\footnotesize Begin forward pass}
  $\Theta_{form}^*(0) = \theta_0$\;
  \For{$t = 1, ..., T$}  { 
   $N = \text{neighbors}(\Theta_{form}^*(t-1))$; \Comment{\scriptsize connected cells}
   $\Theta_{form}^*(t) = \argminprob{N} V_{t}$;
  }
  \textbf{return} $\Theta_{form}^*$
\end{algorithm}

\begin{figure}[h!]
    \includegraphics[width=\linewidth]{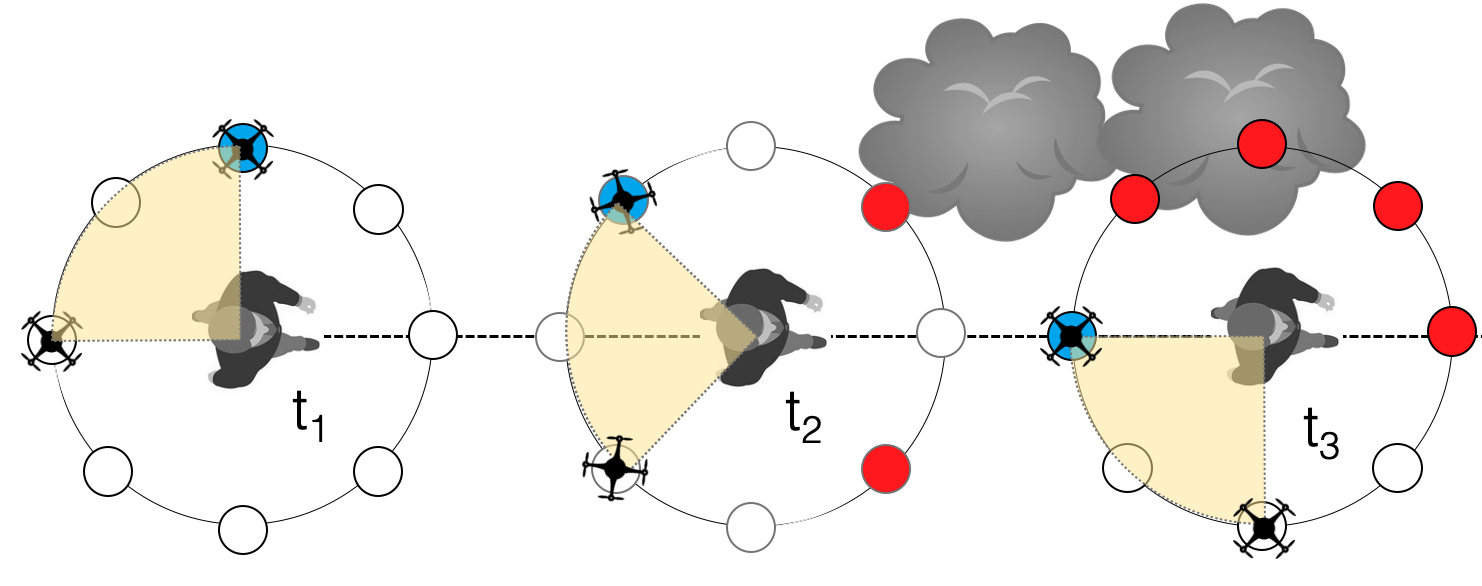}
     \caption{\small Centralized formation planning where formation rotates counter-clockwise to avoid trees. Formation cost map is updated for all time steps, with red as high cost. We apply dynamic programming to solve for the full formation yaw sequence $\Theta_{form}^*$, shown in blue. 
     }
\end{figure}

\subsection{Decentralized Trajectory Optimization}
\label{subsec:decentralized}
After calculating the formation angles $\Theta_{form}^*$ using the centralized planner, we optimize and smoothen individual UAV trajectories at a finer time discretization.
This second step is computed on each UAV's local computer using a decentralized planner. 
While the original waypoints were spaced every $2$ seconds over a $10$-second horizon, here we achieve finer resolutions with $0.5$ s granularity in local planning.
We use the local planner described in \cite{bonatti2020autonomous}, which uses covariant gradient descent to produce locally optimal trajectories while again considering the costs of smoothness, obstacle and occlusions avoidance, and desired formation position of each UAV individually.
In addition, each local planner receives the expected waypoints of all remaining vehicles, and avoids positioning its trajectory within $3m$ of other UAVs.
We run the local planner at $5$ Hz, and use a PID controller for trajectory tracking at $50$ Hz.

\subsection{Offline Skeletal Reconstruction}
\label{subsec:reconstruction}
Once camera images from all UAVs are collected, we post-process the data in an offline phase to generate a sequence of 3D target skeleton poses.
We use AlphaPose, a human skeletal keypoint detector \cite{fang2017rmpe, li2018crowdpose,xiu2018poseflow}, to extract 2D body keypoints from each image.
Next, we linearly triangulate each keypoint using each robot's camera pose and image coordinates to obtain for the keypoint's location in world frame.
\section{Experiments} 
\label{sec:experiments}
Here we detail the simulated and real-world experiments to validate our multi-UAV 3D motion capture system.

\subsection{Simulation Experiments}

\begin{figure}[]
    \includegraphics[width=1.0\linewidth]{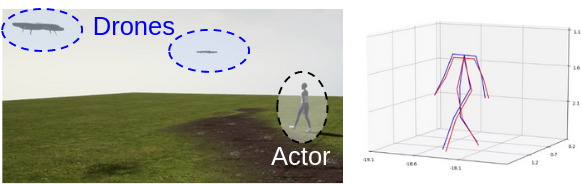}
     \caption{\small Experimental Setup: \textit{(Left)} Photo-realistic simulator with an animated character and drone formation. \textit{(Right)} Ground truth skeleton extracted from simulator (blue) and estimated skeleton with our proposed system (red).}
     \label{fig:sim-setup}
\end{figure}

\textbf{Experimental Setup:}
We quantitatively evaluate our proposed system in a photo-realistic environment, Microsoft AirSim \cite{shah2018airsim} via a custom ROS~\cite{ros} interface, and directly query ground-truth skeleton points from Unreal Engine.
Fig. \ref{fig:sim-setup} shows our simulated experimental setup. 

\begin{figure}[h]
    \centering
    \begin{subfigure}[b]{0.9\textwidth}
        \includegraphics[width=1.0\linewidth]{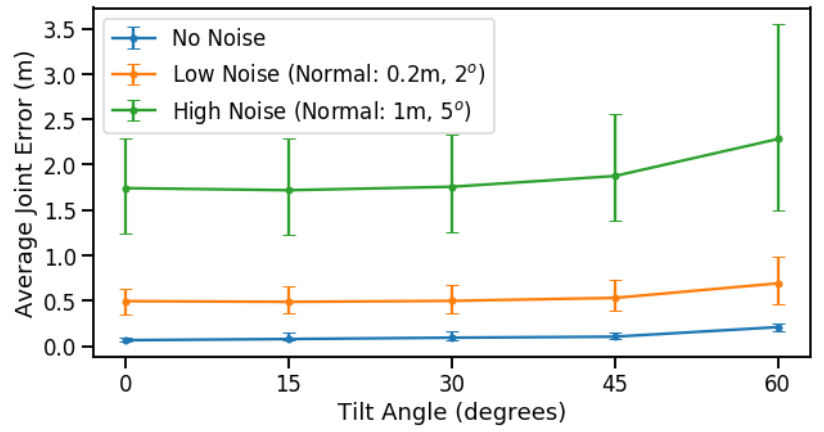}
        \caption{Reconstruction Error vs. Formation Tilt Angle}
        \label{fig:sim-tilt}
    \end{subfigure}
    \hfill
    \begin{subfigure}[b]{0.9\textwidth}
        \centering
        \includegraphics[width=\textwidth]{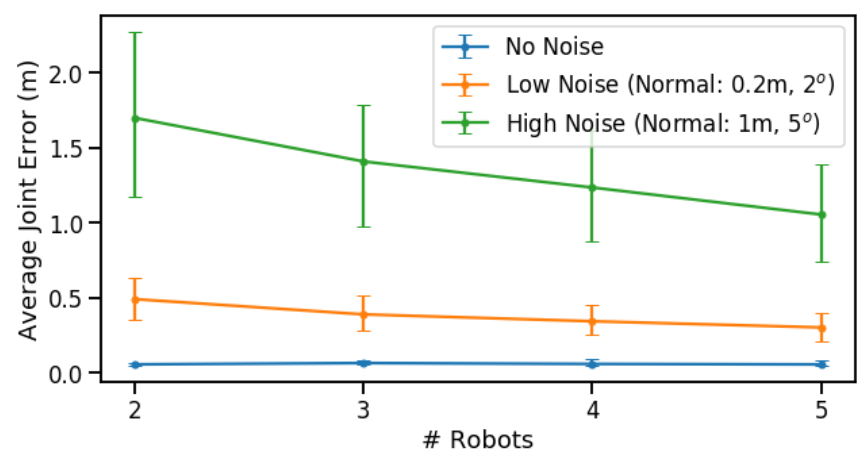}
        \caption{Reconstruction Error vs. \# Robots}
        \label{fig:sim-multi}
    \end{subfigure}
       \caption{\small \textit{a)} Reconstruction Quality vs. Tilt Angle: We observe better reconstruction and resistance to noise with lower tilt angle, with an angle of $0^o$ to $30^o$ giving similar performance. \textit{b)} Reconstruction Quality vs. Number of Robots: With no noise, performance is comparable from $n=2$ to $5$. We observe better resistance to noise as number of robots increases, with marginally decreasing benefit.}
       \label{fig:three graphs}
\end{figure}
\textbf{Sim E1) Reconstruction quality across tilt angles:}
Our first experiment's objective is to quantify the benefit of maintaining low formation tilt angle $\phi_{des}$ for human reconstruction.
To do so, we generated 75 seconds of data of an actor walking with two drones at each tilt angle between $0^o$ to $60^o$ at increments of $15^o$.
Fig. \ref{fig:sim-tilt} shows lower tilt angle yields better reconstruction.
This is expected because, firstly at a high tilt angle, most parts of the target's body are occluded and a slight error in image coordinate can result in large 3D reconstruction error.
Secondly, it is more likely for the keypoint detector to misidentify limbs at a higher angle, possibly due to lack of training data at such angles.
As our camera pose is often noisy for our \textit{in-the-wild} reconstruction application, we also examine how susceptible each tilt angle is to camera pose noise.
Within the same noise level, we observe tilt angles of $0^o$ to $30^o$ provide comparable reconstruction quality.

\textbf{Sim E2) Reconstruction quality using more robots:}
Next, we examine the marginal benefits that more robots bring to reconstruction accuracy. We record 75 seconds of a target walking with with $n=[2,3,4,5]$ drones at a formation tilt angle $\phi_{des}=15^o$. Figure \ref{fig:sim-multi} shows that with no noise, average error is near 0 for all configurations, with a slight decrease at 5 drones. As expected, increasing the number of simultaneous viewpoints helps significantly with noisy camera poses, with decreasing marginal benefits as number of drones increase. At high noise level, the 5-drone configuration reduces error by $\sim30\%$ from the two-drone setup. 
\subsection{Real-World Experiments} 
\begin{figure}[H]
    \centering
    \includegraphics[width=0.7\textwidth]{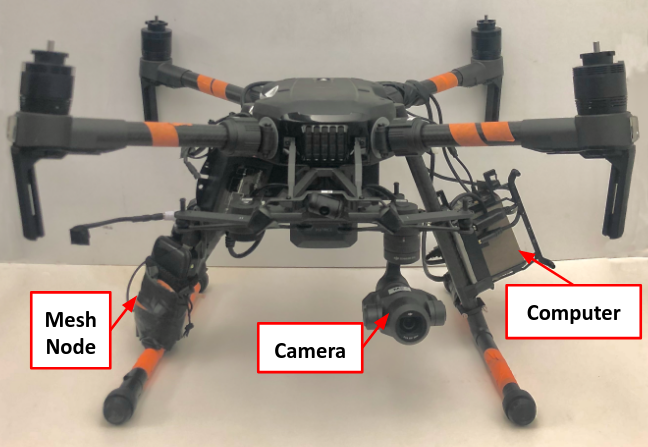}
    \caption{\small System hardware: DJI M210 drone, Intel NUC computer, Ubiquiti mesh nodes and Zenmuse X4S camera gimbal. 
    } 
    \label{fig:exp_platform}
\end{figure}
\textbf{Experimental Setup:}
For real world experiments, we used two DJI M210 drones, one shown in Figure \ref{fig:exp_platform}.
We subsequently refer to these as \emph{drone 1} and \emph{drone 2}.
All processing is done onboard an Intel NUC with 8GB of RAM and an Intel Core i7-8550U processor. Drones communicate with each other with a Ubiquiti WiFi mesh access point \cite{ubiquiti} via the Data Distribution Service networking middleware \cite{pardo2003omgdds}. The leader drone (\textit{drone 1}) runs the centralized planner and sends estimated actor odometry and formation trajectory to \textit{drone 2} for local decentralized planning. Both drones share current odometry and final optimized trajectory for safety. 

An independently controlled DJI Zenmuse X4S gimbal camera records footage. Video frames are processed using an off-the-shelf Intel's OpenVINO MobileNetV2-like pedestrian detector for actor detection.

Our centralized formation planner solves a $10s$ horizon plan with $D=8$ possible discrete formation yaws. The planner runs at $10$Hz with a computation time of $\sim 1.3ms$.

We conduct real-world experiments in a pre-mapped outdoors test site. Range is at $\rho_{des}=10m$  for safety and formation tilt is set at $\phi_{des}=15^o$, which from our simulated results renders good reconstruction.

\begin{figure}[t]
    \centering
    \includegraphics[width=1.0\textwidth]{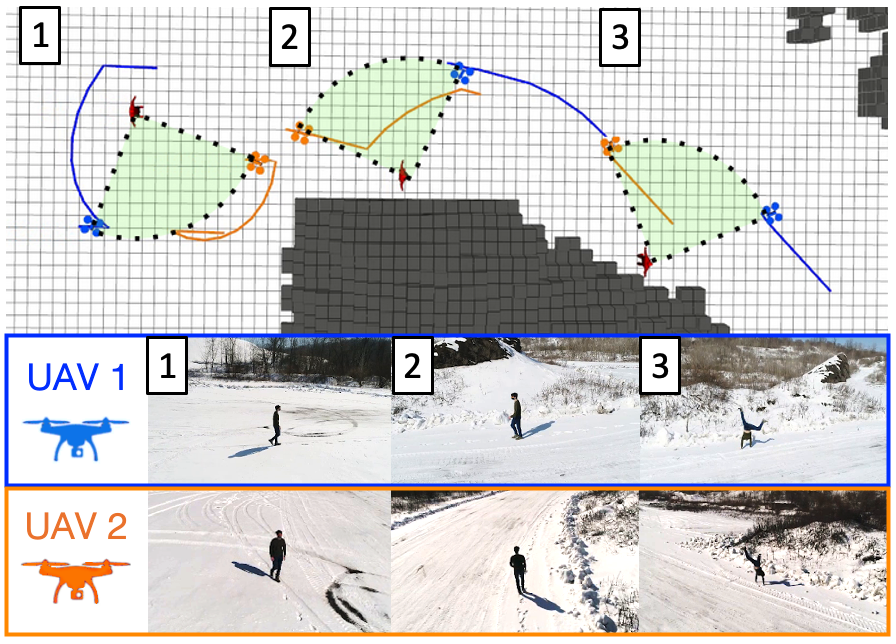}
    \caption{\small Real-life flight among obstacle. Our adaptive formation rotates clockwise avoiding the mound to maintain $90^o$ from each other and a low tilt angle to actor for optimal reconstruction.
    \vspace{-2.0mm}}
    \label{fig:rotate-trial}
\end{figure}

\textbf{Real E1) Formation obstacle avoidance: } 
We  tested  the  system  by recording a moving actor with two drones in a pre-mapped environment.
Fig. \ref{fig:rotate-trial} shows an example trial of our proposed system where the two-drone formation rotated clockwise to avoid colliding with the mound.
The drones are therefore able to maintain a low viewing angle while keeping safe. While the central planner runs at 10Hz, we show three representative timesteps of the experiment for clarity.
The central planner's output at keyframes for \textit{drones} \textit{1} and \textit{2} are colored in blue and  orange respectively. Each UAV then optimizes the coarse formation path with its own local planner for a final smooth, obstacle-free trajectory.
\begin{figure}[h]
    \centering
    \includegraphics[width=1\textwidth]{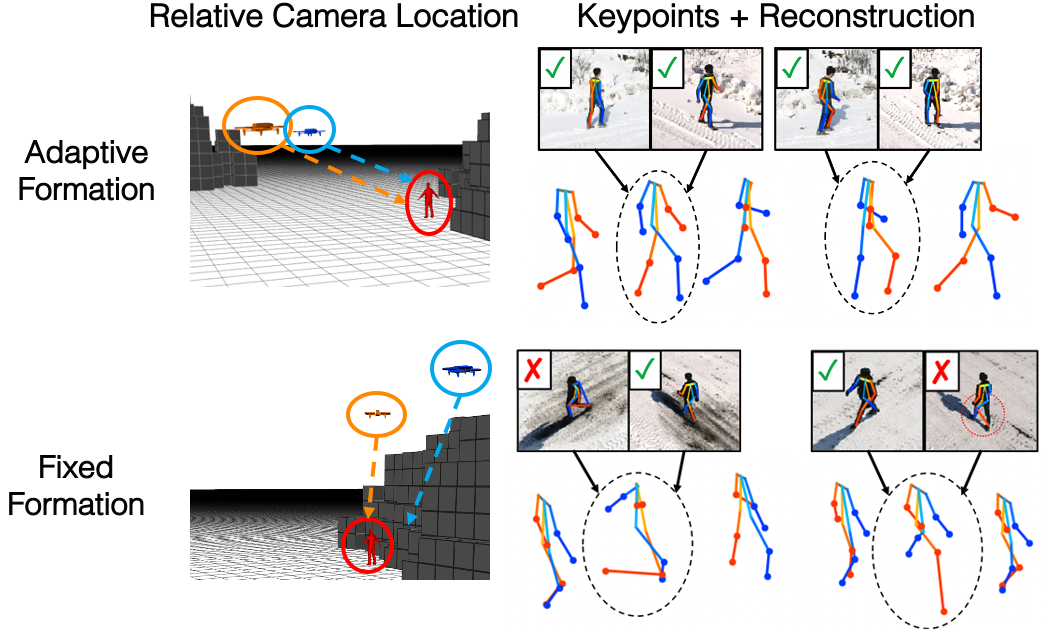}
    \caption{
        \small Reconstruction comparison with and without adaptive formation planning. The formation planning keeps camera at a low tilt angle and significantly improves reconstruction. Without formation planning, the UAVs goes upward to avoid mound. The skeletal keypoint detection fails often for footage collected with fixed formation, resulting in poor reconstruction.
    \vspace{-2.0mm}}
    \label{fig:rw-fixed-compare}
\end{figure}
\textbf{Real E2) Adaptive versus fixed formation: } For the same initial formation angle, we compare the reconstruction performance with and without our adaptive formation planning. 
In the \textit{fixed} trial, the two drones go upward and deviate from the desired tilt angle to avoid mound, resulting in a tilt angle of $\sim60^o$. The high tilt angle results in highly inconsistent reconstruction, due to the increased likelihood of keypoint detection error with examples circled in Fig. \ref{fig:rw-fixed-compare}.
Our proposed adaptive formation planning keeps drones at a low tilt angle, significantly improving the reconstruction quality while avoiding obstacles.

\textbf{Real E3) Reconstructing highly dynamic targets: }
We evaluate the robustness of our proposed system by reconstructing an actor performing abrupt motion changes and highly dynamic movement: jogging and playing soccer.
Figure \ref{fig:rw-dynamic} shows the reconstruction of an actor playing soccer.
As seen in the supplementary video, both UAVs were able to maintain view of the target while keeping desired reconstruction formation angle.
Inlet images in Figure~\ref{fig:rw-dynamic} show the reprojection of the target joints on the UAV's images, with reprojected joints close to actual position.

 \begin{figure}
\includegraphics[width=1\textwidth]{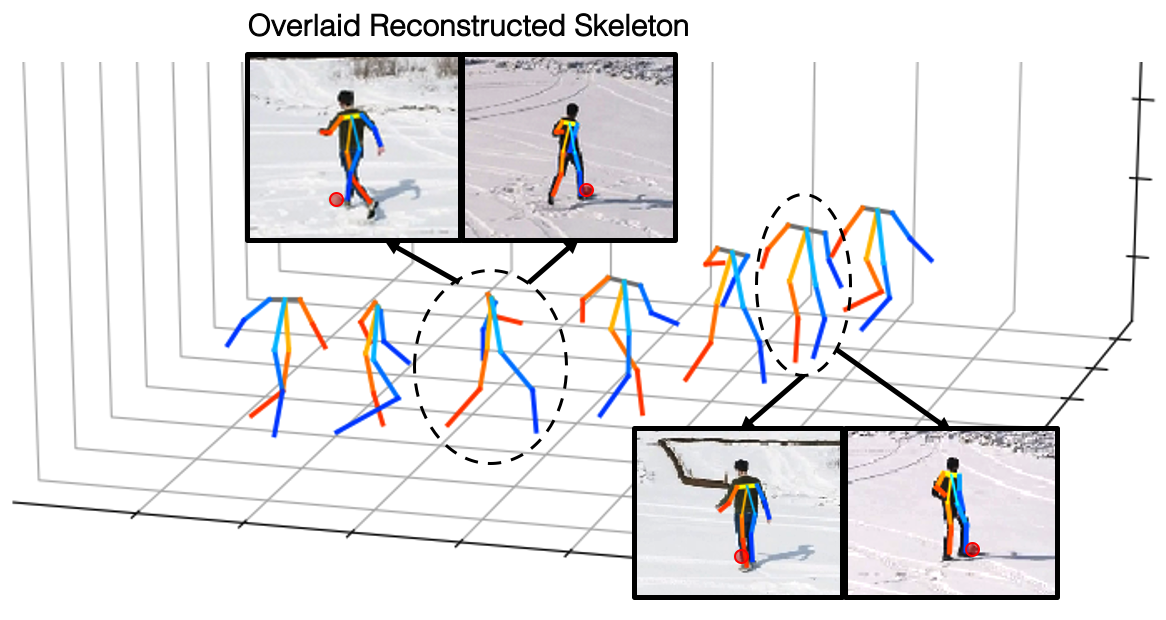}
\caption{\small 3D reconstruction of a highly dynamic real-life actor playing soccer. Our system is able to keep up with abrupt motion changes and provide good reconstruction. Inlet figures show the reprojection of reconstructed 3D skeleton overlaid onto UAV images.} 
\label{fig:rw-dynamic}
\end{figure}

\section{Conclusion and Discussion}
\label{sec:conclusion}
In this paper, we present a collaborative multi-UAV system for 3D human pose reconstruction \textit{in the wild}.
First, we develop a multi-camera coordination scheme to maximize 3D reconstruction quality of dynamic targets while avoiding obstacles and occlusions.
Our approach consists of two steps: 1) a centralized formation planner to compute best camera formation and 2) a decentralized trajectory optimizer to calculate smooth trajectories.
We validate our system in simulated and real-world experiments, and show that it successfully reconstructs targets performing dynamic activities, such as jogging and playing soccer.
Additionally, we provide insights into how reconstruction quality changes with our system under different operating conditions, such as number of drones, camera pose uncertainty and tilt angles.

We find multiple directions for future work in our multi-UAV system.
We are actively working towards the goal of capturing high-fidelity 4D reconstruction of groups of actors and animals in their natural settings, 
with research thrusts in onboard multi-actor detection and tracking \cite{Wang2020_GNNDetTrk, Weng2020_GNNTrkForecast_eccvw},
adaptive multi-agent role reconfiguration to maintain visibility of actors when group splits \cite{engin2020active},
and human mesh reconstruction \cite{jafarian2021tiktok}.





\section*{ACKNOWLEDGMENT}
The authors thank Arthur Bucker, Andrew Ashley and Sam Triest for their help in field experiments. 
CH is supported by the Croucher Foundation. 
This work is supported by the National Science Foundation under grant no. 2024173.

\footnotesize{
\bibliographystyle{IEEEtran}
\bibliography{root}
}
\balance

\end{document}